\documentclass{article}
\usepackage{spconf,amsmath,graphicx}
\usepackage{mwe} 
\usepackage{bm}
\usepackage[T1]{fontenc}
\usepackage{verbatim}
\usepackage{enumitem}
\usepackage{multirow}
\usepackage{booktabs}
\usepackage{environ}
\usepackage{amssymb,amsfonts}
\usepackage{algorithmic}
\usepackage{textcomp}
\usepackage{hyperref}
\setlist{nosep, leftmargin=14pt}

\title{Self-Supervised Slice-to-Volume Reconstruction with Gaussian Representations for Fetal MRI}
\name{\begin{tabular}{c}
Yinsong Wang$^{1}$ \qquad Thomas Fletcher$^{1}$ \qquad Xinzhe Luo$^{1}$
Aine Travers Dineen$^{2}$ \\ \qquad Rhodri Cusack$^{2}$ \qquad Chen Qin$^{1}$
\end{tabular}}
\address{$^{1}$ Department of Electrical and Electronic Engineering and I-X, Imperial College London, London, UK \\
$^{2}$ Trinity College Institute of Neuroscience, Trinity College Dublin}
\begin{document}
\maketitle
\begin{abstract}
Reconstructing 3D fetal MR volumes from motion-corrupted stacks of 2D slices is a crucial and challenging task. Conventional slice-to-volume reconstruction (SVR) methods are time-consuming and require multiple orthogonal stacks for reconstruction. While learning-based SVR approaches have significantly reduced the time required at the inference stage, they heavily rely on ground truth information for training, which is inaccessible in practice. To address these challenges, we propose GaussianSVR, a self-supervised framework for slice-to-volume reconstruction. GaussianSVR represents the target volume using 3D Gaussian representations to achieve high-fidelity reconstruction. It leverages a simulated forward slice acquisition model to enable self-supervised training, alleviating the need for ground-truth volumes.
Furthermore, to enhance both accuracy and efficiency, we introduce a multi-resolution training strategy that jointly optimizes Gaussian parameters and spatial transformations across different resolution levels. Experiments show that GaussianSVR outperforms the baseline methods on fetal MR volumetric reconstruction. Code is available at \href{https://github.com/Yinsong0510/GaussianSVR-Self-Supervised-Slice-to-Volume-Reconstruction-with-Gaussian-Representations}{https://github.com/Yinsong0510/GaussianSVR-Self-Supervised-Slice-to-Volume-Reconstruction-with-Gaussian-Representations}.
\end{abstract}
\begin{keywords}
Slice-to-volume reconstruction, Gaussian Representation, self-supervised learning
\end{keywords}
\section{Introduction}
High-resolution 3D fetal MRI is essential for advancing the understanding of fetal brain development \cite{ benkarim2020novel}; however, it remains highly vulnerable to artifacts resulting from rapid and unpredictable fetal motion. To address this issue, two-dimensional (2D) MRI techniques such as half-Fourier acquisition single-shot fast spin echo (SSFSE) \cite{saleem2014fetal} are used to acquire 2D slices within fractions of a second, effectively freezing in-plane motion. However, residual inter-slice motion and the use of thick slices to preserve signal-to-noise ratio (SNR) limit the ability of cross-sectional views to accurately capture the 3D brain structure, necessitating slice-to-volume reconstruction (SVR) to recover the underlying volume. 

Conventional optimization-based slice-to-volume reconstruction (SVR) methods formulate the volumetric reconstruction as a joint registration and super-resolution problem. Rousseau et al. \cite{rousseau2006registration} and Gholipour et al \cite{gholipour2010robust} reconstruct a high-resolution volume by iteratively estimating slice-to-volume transformations and voxel intensities under a slice acquisition model. Despite their effectiveness, the discretized voxel grid representation makes the complexity and memory footprint of SVR proportional to the number of voxels in the volume. Recently, several works have been proposed to use neural networks for SVR to alleviate the computational burden. Hou et al. \cite{hou20183} proposed a 3D CNN based on pre-trained VGG-16 convolutional layers, with a densely connected head to predict anchor points for individual slices. Xu et al. \cite{xu2022svort} introduced an iterative transformer to jointly estimate transformations and reconstruct the 3D volume. However, these methods require ground-truth transformations for training, which are inaccessible in practice. Subsequently, Xu et al. \cite{10015091} utilized implicit neural representations (INR) to reconstruct the underlying volume, enabling a continuous and resolution-agnostic representation. However, the globally parameterized INR limits its capacity to adapt to local structural variability, resulting in suboptimal reconstruction performance.

To address the limitations, we propose to leverage 3D Gaussian representations to model the underlying 3D volumes. 3D Gaussian Splatting (3DGS) \cite{kerbl20233d} emerged as a ground-breaking technique for novel-view synthesis due to its rapid rasterization and superior rendering quality in comparison to INR. It has also recently raised great interest in the medical imaging domain, including applications on CT reconstruction \cite{li20253dgr}, surgical navigation \cite{fehrentz2025bridgesplat}, and surgical scene reconstruction \cite{paonim2025endoplanar}. However, the application of Gaussian representations to volumetric reconstruction and motion correction has not been explored in prior work. In this work, we present \textbf{GaussianSVR}, a self-supervised SVR framework based on Gaussian representations. Compared to the INR-based method, 3D Gaussian kernels offer spatially localized and independent primitives, enabling fine-grained adaptation to complex anatomical structure while preserving global consistency. Moreover, 3D Gaussian representations have an implicit regularization for the reconstructed volume due to the smooth nature of Gaussian kernels. To better handle Gaussian and motion estimation, we propose a multi-resolution training strategy that performs joint optimization of both parameters hierarchically across multiple resolution levels, as rigid motion can be more reliably estimated at coarser resolutions. Furthermore, we employ a simulated forward slice acquisition model to generate reconstructed stacks from the reconstructed volume and compute their discrepancy with the acquired slices, enabling self-supervised training. Our main contributions are summarized as follows:
\begin{itemize}
    \item We are the first to propose the novel SVR framework based on 3D Gaussian representation.
    \item We introduce a self-supervised multi-resolution training strategy for joint optimization of Gaussian and motion parameters with a simulated slice acquisition model.
    \item Experiments show that GaussianSVR achieves superior reconstruction performance compared to the baseline methods.
\end{itemize}

\section{Methodology}

\begin{figure}[!t]
	\centering
	\includegraphics[width=0.5 \textwidth]{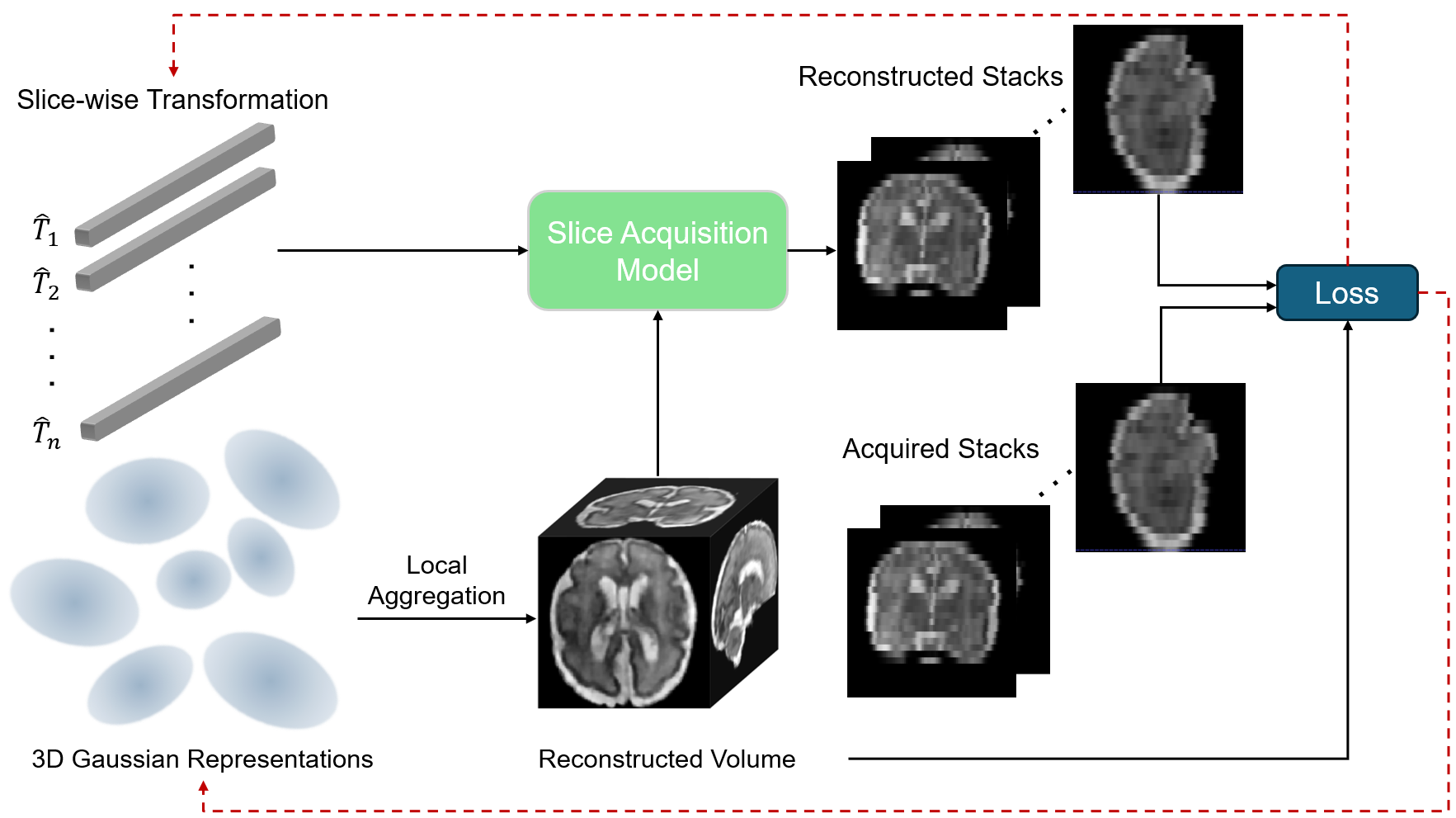} 
	\caption{Overview of the framework of the proposed GaussianSVR. Solid lines indicate forward propagation; dashed lines indicate backward propagation.} 
	\label{figure 2.1}
\end{figure}

Given acquired stacks of 2D slices $\bm{y} = [y_1, \ldots, y_n]$, the objective of slice-to-volume reconstruction (SVR) is to recover the underlying 3D volume $\hat{\bm{x}}$. In the proposed method, $\hat{\bm{x}}$ is represented as a set of 3D Gaussian primitives. Specifically, our framework employs a simulated forward slice acquisition model to generate reconstructed stacks based on the estimated slice-wise transformations, thereby enabling self-supervised joint optimization of the Gaussian and transformation parameters through a multi-resolution training strategy. The overall framework is illustrated in Figure~\ref{figure 2.1}.

\subsection{Volumetric Reconstruction by 3D Gaussian Representations}\label{sec 2.1}

3D Gaussian Splatting (3DGS) \cite{kerbl20233d} represents a 3D scene as a set of Gaussian primitives ${G_j \mid j=1,\dots,J}$. Each Gaussian $G_j$ is parameterized by its center $\mu_j$, covariance $\Sigma_j$, opacity $\alpha_j$, and spherical harmonic coefficients $SH_j$ for view-dependent appearance. Formally, a 3D scene can be expressed as $G = \left\{G_j : \mu_j, \Sigma_j, \alpha_j, SH_j|1,...,J\right\}$. To ensure the covariance matrix $\Sigma_j$ remains positive semi-definite, it is decomposed into a rotation matrix $R_j$ and a diagonal scaling matrix $S_j$ as: $\Sigma_j = R_jS_j^2R_j^T$. A 3D Gaussian primitive in continuous space is then defined as:
\begin{equation}
	G_j(\mathbf{x})=e^{-\frac{1}{2}(\mathbf{x}-\mu_j)^T \Sigma^{-1}_j(\mathbf{x}-\mu_j)},
\end{equation}
where $\mathbf{x}$ denotes a spatial location in 3D space. Rendering a novel view of the scene involves projecting these 3D Gaussians onto the 2D image plane.

While the original 3DGS framework was developed for natural images, its appearance-related parameters, opacity $\alpha_j$ and spherical harmonics $SH_j$, are inapplicable to medical imaging. Following Li et al. \cite{li20253dgr}, we therefore remove these parameters and introduce an intensity coefficient $I_j$ to represent the MRI intensity value at each Gaussian center. The 3D MRI volume is thus represented as $G = \left\{G_j : \mu_j, \Sigma_j, I_j|1,...,J\right\}$, and the contribution of each Gaussian to a spatial point $\mathbf{x}$ is formulated as
\begin{equation}
	G_j(\mathbf{x}| \mu_j, \Sigma_j, I_j)= I_je^{-\frac{1}{2}(\mathbf{x}-\mu_j)^T \Sigma^{-1}_j(\mathbf{x}-\mu_j)}.
\end{equation}
The covariance matrix $\Sigma_j$ is parameterized in the same way as in the original 3DGS \cite{kerbl20233d} by a scaling matrix $S_j$ and a rotation matrix $R_j$, such that $\Sigma_j = R_jS_j^2R_j^T$ to ensure the positive semi-definiteness of $\Sigma_j$ and allow for independent optimization of these parameters.

To optimize the parameters of our 3D Gaussian representations and accurately estimate the MRI intensity values of the underlying 3D volume, we modify the standard rendering process used in original 3DGS \cite{kerbl20233d}. Instead of employing the conventional splatting-based rendering, we compute the volumetric intensity at any spatial location $x$ through a localized aggregation of contributions from neighboring Gaussians. To improve computational efficiency while maintaining reconstruction fidelity, we restrict the computation within a 99\% ($\mu_j\pm3\sigma_j$) confidence interval for each Gaussian $G_j$. Therefore, the MRI intensity value $V(\mathbf{x})$ for any give point $x$ of the reconstructed 3D volume can be formulated as,
\begin{equation}
    V(\mathbf{x}|\mu_j, \Sigma_j, I_j)= \sum_{j:||x-\mu_j||\leq3\sigma_j}^J G_j(\mathbf{x}|\mu_j, \Sigma_j, I_j).
\end{equation}

In this formulation, a 3D MR volume is represented as a set of Gaussian primitives, which allows spatially localized independent optimization for each primitive, enabling fine-grained adaptation for complex anatomical structures.

\subsection{Slice Acquisition Model}\label{sec 2.2}
Given the underlying 3D volume $x$, the acquired stacks of 2D slices $\bm{y} = [y_1,..., y_n]$, can be obtained by the forward slice acquisition model, which is formulated as follows \cite{xu2022svort},
\begin{equation}\label{SAM}
    y_i = \bm{D}\bm{B}\bm{T}_i\bm{x}; \; \; i=1,...,n.
\end{equation}
Here $\bm{y}_i$ represents the $i$ th 2D slice, $n$ is the total number of slices, $\bm{T}_i$ is the slice-wise transformation parameters of $i$ th 2D slice, which describe the rotations and translations of $i$ th plane within a canonical 3D atlas space. $\bm{B}$ represents the Point-Spread-Function (PSF) blurring matrix of the MRI signal acquisition process, and $\bm{D}$ is a down-sampling matrix. In this work, we modeled the PSF as an anisotropic 3D Gaussian distribution. Based on this, the reconstructed stack of slices $\bm{\hat{y}} = [\hat{y}_1,..., \hat{y}_n]$ thus can be obtained using the reconstructed 3D volume $\bm{\hat{x}}$ through
\cite{xu2022svort},
\begin{equation}\label{SAM}
    \hat{y}_i = \bm{D}\bm{B}\bm{\hat{T}}_i\bm{\hat{x}}; \; \; i=1,...,n,
\end{equation}
where $\bm{\hat{T}}_i$ represents the optimized transformation parameters of $i$ th slice from GaussianSVR.

\subsection{Training}
To improve both accuracy and efficiency, we propose a multi-resolution training strategy to jointly optimize slice-wise transformations and the parameters of Gaussian representations across different spatial resolutions. The training process is formulated in a self-supervised manner without reliance on ground-truth supervision.

\noindent\textbf{Low-Resolution Optimization.} In the first stage, both the slice-wise transformations and Gaussian parameters are optimized at a coarse spatial resolution. Operating at low resolution stabilizes training and accelerates convergence, as the rigid slice-wise motion patterns are easier to capture when fine structural details are suppressed. This stage yields a robust initialization for the subsequent refinement phase.

\noindent\textbf{High-Resolution Optimization.} The second stage refines both Gaussian and transformation parameters obtained from the coarse-level optimization. By increasing the spatial resolution, GaussianSVR recovers fine-grained anatomical details while further improving alignment accuracy. This coarse-to-fine strategy promotes a well-conditioned optimization landscape and mitigates convergence to local minima.

\noindent\textbf{Self-supervised training via Slice Acquisition Model.} An overview of the framework is illustrated in Figure~\ref{figure 2.1}. To enable self-supervised learning, the reconstructed 3D Gaussian volume is projected back into 2D slice space using the forward slice acquisition model described in Section~\ref{sec 2.2}. Given the estimated transformations $\bm{\hat{T}}$ and the simulated point-spread function (PSF) in Section~\ref{sec 2.2}, the reconstructed stacks of slices $\hat{\bm{y}} = [\hat{y}_1, \ldots, \hat{y}_n]$ can then be obtained and compared with the acquired stacks $\bm{y} = [y_1, \ldots, y_n]$ to formulate the self-supervision. The Gaussian and transformation parameters are then jointly optimized by minimizing a reconstruction loss that combines an $\mathcal{L}_1$ data fidelity term, a differentiable structural similarity ($D$-SSIM) term, and a total variation (TV) regularization term to promote spatial smoothness in the reconstructed volume. The overall loss function is defined as:
\begin{equation}
\mathcal{L} = \tfrac{1}{n}\!\sum_{i=1}^n \!\big(\|y_i-\hat{y}_i\|_1 + \lambda_1 D\text{-SSIM}(y_i,\hat{y}_i)\big) + \lambda_2 TV(\hat{\bm{x}}),
\end{equation}
where $TV(\hat{\bm{x}})$ denotes the total variation regularization applied to the reconstructed volume $\hat{\bm{x}}$, and $\lambda_1$ and $\lambda_2$ denotes the respectively hyperparameter for $D$-SSIM and TV loss. 
This self-supervised formulation enables GaussianSVR to jointly optimize both transformation and Gaussian parameters directly from raw slice acquisitions, ensuring robust and high-fidelity volumetric reconstruction.

\section{Experiments}
\noindent\textbf{Dataset.} We evaluate our proposed GaussianSVR on the Fetal Tissue Annotation Challenge (FeTA) dataset \cite{payette2021automatic},  which consists of T2-weighted (T2w) fetal brain MR images. We randomly selected 30 volumes as ground truths for evaluation. The volumes are registered to a fetal brain atlas \cite{DVN/WE9JVR_2023}, and resampled to the resolution of $0.8\times0.8\times0.8$ mm. We simulate 2D slices with a resolution of 1 mm $ \times 1$ mm, slice thickness between 2.5 and 3.5 mm, and size of $128 \times 128$. For each subject, three image stacks comprising 15-30 slices were simulated along orthogonal orientations, following ~\cite{xu2022svort}, with fetal brain motion trajectories generated following ~\cite{xu2021stress}.

\noindent\textbf{Evaluation Metrics.}
To evaluate the reconstruction accuracy with respect to the ground-truth volume, we assessed the reconstructed results using peak signal-to-noise ratio (PSNR), structural similarity index (SSIM) \cite{1284395}, and normalized root mean square error (NRMSE).

\noindent\textbf{Baseline Methods.}
GaussianSVR is compared with three representative slice-to-volume reconstruction methods: the conventional NiftyMIC \cite{ebner2020automated}, the transformer-based learning approach SVoRT \cite{xu2022svort}, and a recent optimization-based method using implicit neural representation, NeSVoR \cite{10015091}.

\noindent\textbf{Implementation Details.}
 The method was implemented in PyTorch and trained on an NVIDIA A6000 Ada GPU using the Adam optimizer \cite{adam2014method}. The transformation parameters estimated by the pretrained SVoRT model \cite{xu2022svort} are employed as initialization for GaussianSVR. For multi-resolution training, the volume is reconstructed at a resolution downsampled by a factor of two during the low-resolution stage, and refined at full resolution during the high-resolution stage. For Gaussian parameters, the learning rate of the mean $\mu$ decayed from $2\times10^{-3}$ to $2\times10^{-6}$, while constant rates of $0.05$, $0.005$, and $0.001$ were used for intensity $I$, scaling $S$, and rotation $R$, respectively. For transformation parameters, learning rates were set to $5\times10^{-4}$ for translation and $5\times10^{-5}$ for rotation.

\begin{figure}[!t]
	\centering
	\includegraphics[width=0.48 \textwidth]{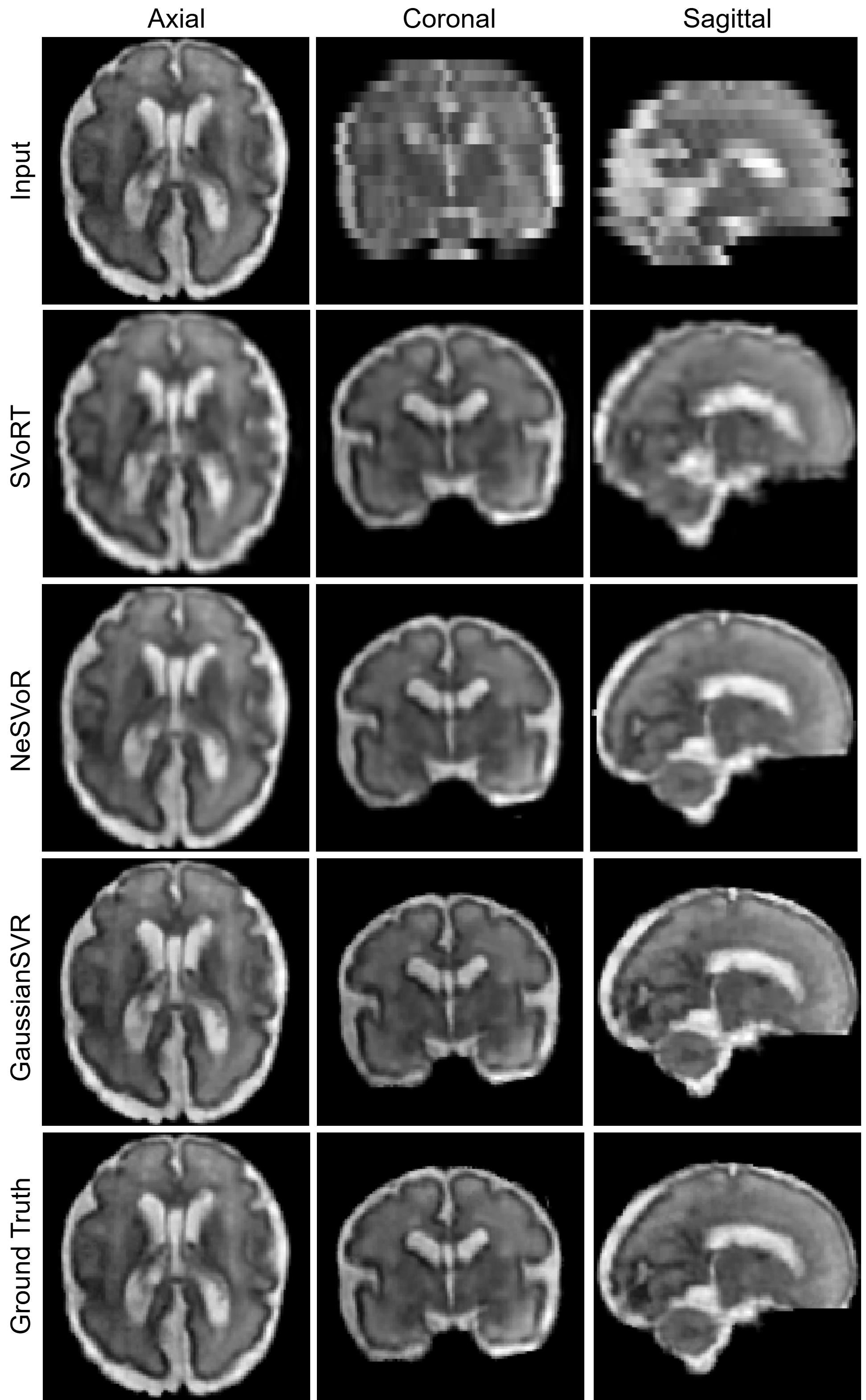} 
	\caption{Qualitative volumetric reconstruction results on a single subject from motion-corrupted scans on the FeTA dataset.} 
	\label{figure 4.1}
\end{figure}

\section{Results and Discussion}
\label{sec:R&D}
\noindent\textbf{Comparison studies.}
Table \ref{table_1} reports the quantitative volumetric reconstruction performance of 3D fetal brain MRI on the FeTA dataset. We report the mean and standard deviation of the reconstruction results of the 30 test subjects. It can be observed that GaussianSVR achieves the highest reconstruction accuracy, achieving a 2.9\% improvement in terms of PNSR compared to the second-best approach, NeSVoR. These results demonstrate that GaussianSVR substantially outperforms existing reconstruction approaches in both structural preservation and intensity consistency. The notably higher SSIM and lower NRMSE indicate that GaussianSVR more effectively maintains anatomical integrity and minimizes reconstruction artifacts. Figure \ref{figure 4.1} shows the qualitative reconstruction results of GaussianSVR compared with other baseline methods. It can be observed that GaussianSVR can reconstruct more fine-grained details. Compared to NeSVoR, GaussianSVR can reconstruct the volume with sharper details, demonstrating that Gaussian representations can achieve high-fidelity volumetric reconstruction.

\begin{table}[!t]
\setlength{\tabcolsep}{1.5 mm}
\caption{Quantitative Results on FeTA datasets. * represents GaussianSVR significantly outperformed with $p$-value $< 0.01$ in a paired $t$-test. (Standard deviation in parentheses)}
\centering
\scalebox{0.80}{
\begin{tabular}{ccccccc}
  \toprule[1pt]  
   \textbf{Methods} & PSNR / dB $\uparrow$ & SSIM $\uparrow$ & NRMSE $\downarrow$ \\
	\cmidrule(lr){1-4}
 NiftyMIC & 21.17*(1.95) & 0.7653*(0.0559) & 0.0989*(0.0234)\\
    SVoRT & 23.98*(2.65) & 0.8209*(0.0618) & 0.0905*(0.1227)\\
    NeSVoR & 25.58*(1.81) & 0.8940*(0.0407) & 0.0536(0.0105) \\
    GaussianSVR (Ours) & \textbf{28.19}(3.02) & \textbf{0.9281}(0.0552) & \textbf{0.0468}(0.0219) \\
	\bottomrule[1pt]
\end{tabular}}
\label{table_1}
\end{table}

\noindent\textbf{Ablation Studies.} 
We conducted ablation studies to assess the effectiveness of the proposed multi-resolution and joint optimization strategy. The results are summarized in Table~\ref{table_2}. Removing multi-resolution training degrades model performance, likely because slice-wise transformations are more stable and converge more effectively at lower resolutions. Omitting transformation optimization causes a substantial performance drop, indicating that jointly optimizing transformations helps the model escape local minima and achieve improved global convergence.
\begin{table}[!t]
\setlength{\tabcolsep}{1.5 mm}
\caption{Ablation study on each module of GaussianSVR. * represents GaussianSVR significantly outperformed with $p$-value $< 0.01$ in a paired $t$-test. (Standard deviation in parentheses)}
\centering
\scalebox{0.88}{
\begin{tabular}{ccccccc}
  \toprule[1pt]
   \textbf{Methods} & PSNR / dB $\uparrow$ & SSIM $\uparrow$ \\
	\cmidrule(lr){1-3}
 w/o low resolution & 27.08*(3.89) & 0.9134*(0.0547) \\
    w/o transformation optimization & 22.86*(2.38) & 0.8148*(0.0752) \\
    GaussianSVR (Ours) & \textbf{28.19}(3.02) & \textbf{0.9281}(0.0552) \\
	\bottomrule[1pt]
\end{tabular}}
\label{table_2}
\end{table}

\section{Conclusion}
In this work, we propose GaussianSVR, a self-supervised slice-to-volume reconstruction (SVR) framework based on 3D Gaussian representations. GaussianSVR employs 3D Gaussian kernels to model the volumetric structure, enabling high-fidelity reconstruction through spatially localized and independent primitives that facilitate fine-grained detail reconstruction. A multi-resolution training strategy is proposed to jointly optimize Gaussian and transformation parameters, leading to more accurate motion estimation and reconstruction. Furthermore, a simulated forward slice acquisition model allows self-supervised training by generating reconstructed stacks of slices and comparing them with the acquired stacks of slices. Experimental results demonstrate that GaussianSVR outperforms baseline methods in reconstruction quality. Future work will explore the resolution-agnostic capability of GaussianSVR and its potential for volumetric reconstruction from a single stack of slices.

\section{Acknowledgments}
This work was supported by the Engineering and Physical Sciences Research Council [grant number EP/Y002016/1] and by Research Ireland under FreezeMotion project [grant number 22/FFP-A/11050]. X. Luo was supported by the Engineering and Physical Sciences Research Council [grant number EP/X039277/1].
\label{sec:acknowledgments}

\section{Compliance with Ethical Standards}
This research study utilized publicly available human subject data from the Fetal Tissue Annotation Challenge (syn25649159), for which ethical approval was obtained by the original data collectors as reported in the associated publications.

\bibliographystyle{IEEEbib}
\bibliography{strings,refs}

@article{rousseau2006registration,
  title={Registration-based approach for reconstruction of high-resolution in utero fetal MR brain images},
  author={Rousseau, Francois and Glenn, Orit A and Iordanova, Bistra and Rodriguez-Carranza, Claudia and Vigneron, Daniel B and Barkovich, James A and Studholme, Colin},
  journal={Academic radiology},
  volume={13},
  number={9},
  year={2006},
  publisher={Elsevier}
}

@article{gholipour2010robust,
  title={Robust super-resolution volume reconstruction from slice acquisitions: application to fetal brain MRI},
  author={Gholipour, Ali and Estroff, Judy A and Warfield, Simon K},
  journal={IEEE transactions on medical imaging},
  volume={29},
  number={10},
  pages={1739--1758},
  year={2010},
  publisher={IEEE}
}

@article{kerbl20233d,
  title={3d gaussian splatting for real-time radiance field rendering.},
  author={Kerbl, Bernhard and Kopanas, Georgios and Leimk{\"u}hler, Thomas and Drettakis, George},
  journal={ACM Trans. Graph.},
  volume={42},
  number={4},
  pages={139--1},
  year={2023}
}

@article{payette2021automatic,
  title={An automatic multi-tissue human fetal brain segmentation benchmark using the fetal tissue annotation dataset},
  author={Payette, Kelly and de Dumast, Priscille and Kebiri, Hamza and Ezhov, Ivan and Paetzold, Johannes C and Shit, Suprosanna and Iqbal, Asim and Khan, Romesa and Kottke, Raimund and Grehten, Patrice and others},
  journal={Scientific data},
  volume={8},
  number={1},
  pages={167},
  year={2021},
  publisher={Nature Publishing Group UK London}
}

@article{hou20183,
  title={3-D reconstruction in canonical co-ordinate space from arbitrarily oriented 2-D images},
  author={Hou, Benjamin and Khanal, Bishesh and Alansary, Amir and McDonagh, Steven and Davidson, Alice and Rutherford, Mary and Hajnal, Jo V and Rueckert, Daniel and Glocker, Ben and Kainz, Bernhard},
  journal={IEEE transactions on medical imaging},
  volume={37},
  number={8},
  year={2018},
  publisher={IEEE}
}

@ARTICLE{10015091,
  author={Xu, Junshen and Moyer, Daniel and Gagoski, Borjan and Iglesias, Juan Eugenio and Grant, P. Ellen and Golland, Polina and Adalsteinsson, Elfar},
  journal={IEEE Transactions on Medical Imaging}, 
  title={NeSVoR: Implicit Neural Representation for Slice-to-Volume Reconstruction in MRI}, 
  year={2023},
  volume={42},
  number={6},
  pages={1707-1719},
  doi={10.1109/TMI.2023.3236216}}

@article{saleem2014fetal,
  title={Fetal MRI: An approach to practice: A review},
  author={Saleem, Sahar N},
  journal={Journal of advanced research},
  volume={5},
  number={5},
  pages={507--523},
  year={2014},
  publisher={Elsevier}
}

@inproceedings{xu2022svort,
  title={SVoRT: Iterative transformer for slice-to-volume registration in fetal brain MRI},
  author={Xu, Junshen and Moyer, Daniel and Grant, P Ellen and Golland, Polina and Iglesias, Juan Eugenio and Adalsteinsson, Elfar},
  booktitle={MICCAI},
  year={2022}
}

@article{li20253dgr,
  title={3DGR-CT: Sparse-view CT reconstruction with a 3D Gaussian representation},
  author={Li, Yingtai and Fu, Xueming and Li, Han and Zhao, Shang and Jin, Ruiyang and Zhou, S Kevin},
  journal={Medical Image Analysis},
  pages={103585},
  year={2025},
  publisher={Elsevier}
}

@data{DVN/WE9JVR_2023,
author = {Gholipour, Ali and Velasco-Annis, Clemente and Rollins, Caitlin K. and Vasung, Lana and Ouaalam, Abdelhakim and Ortinau, Cynthia and Akhondi-Asl, Alireza and Clancy, Sean and Yang, Edward and Estroff, Judy and Warfield, Simon K.},
publisher = {Harvard Dataverse},
title = {{IMAGINE Fetal T2-weighted MRI Atlas}},
UNF = {UNF:6:9QqXyI3PtEeSqF3xiW/LHw==},
year = {2023},
version = {V1},
doi = {10.7910/DVN/WE9JVR},
url = {https://doi.org/10.7910/DVN/WE9JVR}
}

@ARTICLE{1284395,
  author={Zhou Wang and Bovik, A.C. and Sheikh, H.R. and Simoncelli, E.P.},
  journal={IEEE Transactions on Image Processing}, 
  title={Image quality assessment: from error visibility to structural similarity}, 
  year={2004},
  volume={13},
  number={4},
  pages={600-612},
  doi={10.1109/TIP.2003.819861}}

@article{ebner2020automated,
  title={An automated framework for localization, segmentation and super-resolution reconstruction of fetal brain MRI},
  author={Ebner, Michael and Wang, Guotai and Li, Wenqi and Aertsen, Michael and Patel, Premal A and Aughwane, Rosalind and Melbourne, Andrew and Doel, Tom and Dymarkowski, Steven and others},
  journal={NeuroImage},
  volume={206},
  pages={116324},
  year={2020},
  publisher={Elsevier}
}

@article{adam2014method,
  title={A method for stochastic optimization},
  author={Adam, Kingma DP Ba J and others},
  journal={arXiv preprint arXiv:1412.6980},
  volume={1412},
  number={6},
  year={2014}
}

@inproceedings{xu2021stress,
  title={STRESS: Super-resolution for dynamic fetal MRI using self-supervised learning},
  author={Xu, Junshen and Abaci Turk, Esra and Grant, P Ellen and Golland, Polina and Adalsteinsson, Elfar},
  booktitle={MICCAI},
  year={2021}
}

@article{benkarim2020novel,
  title={A novel approach to multiple anatomical shape analysis: application to fetal ventriculomegaly},
  author={Benkarim, Oualid and Piella, Gemma and Rekik, Islem and Hahner, Nadine and Eixarch, Elisenda and Shen, Dinggang and Li, Gang and Ballester, Miguel Angel Gonz{\'a}lez and Sanroma, Gerard},
  journal={Medical image analysis},
  volume={64},
  pages={101750},
  year={2020},
  publisher={Elsevier}
}

@inproceedings{fehrentz2025bridgesplat,
  title={BridgeSplat: Bidirectionally Coupled CT and Non-rigid Gaussian Splatting for Deformable Intraoperative Surgical Navigation},
  author={Fehrentz, Maximilian and Winkler, Alexander and Heiliger, Thomas and Haouchine, Nazim and Heiliger, Christian and Navab, Nassir},
  booktitle={MICCAI},
  year={2025}
}

@inproceedings{paonim2025endoplanar,
  title={EndoPlanar: Deformable Planar-Based Gaussian Splatting for Surgical Scene Reconstruction},
  author={Paonim, Thatphum and Sasnarukkit, Chayapon and Nupairoj, Natawut and Vateekul, Peerapon},
  booktitle={MICCAI},
  year={2025}
}

\end{document}